\documentclass{article}

\usepackage[final]{nips_2016}

\usepackage[utf8]{inputenc} % allow utf-8 input
\usepackage[T1]{fontenc}    % use 8-bit T1 fonts
\usepackage{hyperref}       % hyperlinks
\usepackage{url}            % simple URL typesetting
\usepackage{booktabs}       % professional-quality tables
\usepackage{amsfonts}       % blackboard math symbols
\usepackage{nicefrac}       % compact symbols for 1/2, etc.
\usepackage{microtype}      % microtypography
\usepackage{comment}
\usepackage{relsize}
\usepackage{graphicx}
\usepackage{amsfonts, amssymb,amsmath}
\usepackage{array,float}
\usepackage{soul}
\usepackage{subcaption}
\usepackage{wrapfig}

\def \Rm{{\mathbb{R}}}

\def \Abf{{\mathbf A}}

\def \Dbf{{\mathbf D}}

\def \Hbf{{\mathbf H}}

\def \Wbf{{\mathbf W}}
\def \xbf{{\mathbf x}}
\def \Xbf{{\mathbf X}}

\def \zbf{{\mathbf z}}
\def \Zbf{{\mathbf Z}}
\def \Hbf{{\mathbf H}}
\def \0bf{{\mathbf 0}}

\def \alphabf{{\boldsymbol{\alpha}}}

\usepackage{todonotes}

\title{Extraction of Airways using Graph Neural Networks}

\author{
  Raghavendra Selvan \\
  University of Copenhagen\\
  \texttt{raghav@di.ku.dk}
  \And
  Thomas Kipf \\
  University of Amsterdam \\
  \texttt{t.n.kipf@uva.nl}
  \And 
  Max Welling \\
  University of Amsterdam \\
  CIFAR\footnote{Canadian Institute for Advanced Research}\\
  \texttt{m.welling@uva.nl}
  \And 
  Jesper H. Pedersen \\
  University Hospital of Copenhagen\\
   \texttt{jped0106@regionh.dk}
  \And 
  Jens Petersen \\
  University of Copenhagen\\
  \texttt{phup@di.ku.dk}
  \And 
  Marleen de Bruijne \\
  University of Copenhagen \\
  Erasmus MC, Rotterdam\\
  \texttt{marleen@di.ku.dk}
}

\begin{document}

\maketitle
\vspace{-0.5cm}
\begin{abstract}
We present extraction of tree structures, such as airways, from image data as a graph refinement task. To this end, we propose a graph auto-encoder model that uses an encoder based on graph neural networks (GNNs) to learn embeddings from input node features and a decoder to predict connections between nodes. Performance of the GNN model is compared with mean-field networks in their ability to extract airways from 3D chest CT scans. 
\end{abstract}

\section{Introduction}
\label{sec:intro}

Extraction of trees, like airways, from image data is useful in many image analysis applications. In this work, we pose tree extraction as a graph refinement task, of recovering underlying connectivity that corresponds to structures of interest from an over-complete graph. In previous work from the authors~\citep{selvan2018mean}, graph refinement for extraction of tree structures was performed in a probabilistic inference setting using mean-field networks (MFNs) yielding competitive results. 

Graph neural networks (GNNs) \citep{scarselli2009graph,li2015gated} are a new class of models that are interpreted as generalisation of the message passing algorithms such as MFNs~\citep{li2014mean}. These models can be used as trainable end-to-end systems allowing inference using message passing even in cases where closed form algorithms do no exist~\citep{yoon2018inference}. The lack of prior work on using GNNs for graph refinement in medical imaging motivated us to examine the usefulness of GNNs for such tasks and present a preliminary investigation of using a variant of GNN, introduced in~\citep{kipf2016variational} known as Graph Auto-Encoders (GAEs), to extract airways.
%While there are several closely related formulations of GNNs, we focus on a variant introduced in~\citep{kipf2016variational} known as Graph Auto-Encoders (GAEs). 
% * <phup@di.ku.dk> 2018-04-10T07:42:08.014Z:
% 
% > generalisation
% They are a bit more than that, I guess?
% 
% ^ <phup@di.ku.dk> 2018-04-10T07:43:04.851Z.

%In this work, we develop supervised GNN models to perform graph refinement, with a focus on extracting airway trees from CT data.  We pre-process CT images to obtain over-complete graphs with node features following the method in~\citep{selvan2017extraction}. These graphs are input to the GNN model and trained by minimising the loss between the predicted and ground truth adjacency matrices. We evaluate the GNN model on 3D, low-dose, chest scans from the Danish lung cancer screening trial~\citep{pedersen2009danish} using two datasets. The first of which comprises of 32 scans with expert corrected reference segmentations, and an additional 100 scans with automatically generated labels. Due to the limited availability of labelled data ($32$), we use automatically generated reference labels to perform hyperparameter tuning and pre-training of the network. We compare the performance of GNN model with the MFN model, and a second method that uses region growing on probability images~\citep{lo2010vessel} which is similar to one of the best performing methods in EXACT'09 airway extraction challenge~\citep{lo2012extraction}. To the extent of our knowledge, this is the first attempt at using GNNs for graph refinement tasks on medical imaging data.

\section{Method}
\label{sec:meth}

%\subsection{Graph refinement model}

%Given an over-complete input graph, $\Gcal:\{\Ncal,\Ecal\}$, with nodes $i\in\Ncal$, $F$-dimensional node features, $\xbf_i \in \Rm^F$ and edges in $(i,j)\in\Ecal$, we are interested in recovering a subgraph, ${\Gcal^{\prime}}:\{\Ncal^{\prime},\Ecal^{\prime}\}$ by refining the input graph. 
Given an over-complete input graph with $N$ nodes, an adjacency matrix, $\Abf_\text{I} \in \{0,1\}^{N\times N}$, describing connectivity between nodes, and a node feature matrix, $\Xbf \in \Rm^{N\times F}$, where $F$ is dimensionality of the features, we seek a model that can recover the ground truth adjacency matrix, ${\Abf} \in \{0,1\}^{N\times N}$, corresponding to structures of interest from the input graph, i.e. $f(\Abf_\text{I}, \Xbf) \rightarrow \Abf$.
%. That is,   i.e., $f(\Abf_\text{I}, \Xbf) \rightarrow \Abf$, where $\Xbf \in \Rm^{N\times F}$ is the node feature matrix. 
For airway data, we pre-process the CT scans into graph-structured data, following the method in~\citep{selvan2017extraction}, such that each node is  associated with a $7-$dimensional Gaussian density, with local radius, position and orientation in 3D, and their variances, as the node feature vector: $\xbf_i = [\xbf^i_{\mu}, \xbf^i_{\sigma^2}]$, comprising of mean, $\xbf^i_{\mu}\in \Rm^{7\times 1}$, and variance for each feature, $\xbf^i_{\sigma^2}\in \Rm^{7\times 1}$. 

%\subsection{\bf Graph neural networks } 
GAEs are comprised of an encoder that learns an embedding, $\Zbf \in \Rm^{N\times E}$, based on an input adjacency matrix and node features: $\Zbf = f(\Abf_\text{I}, \Xbf)$, where $E$ is the dimensionality of the learnt embedding space. The encoder $f(\cdot)$ comprises of a graph convolution operation and a non-linearity $\mathlarger \sigma(\cdot)$. The encoder used in this work uses a formulation similar to~\citep{kipf2016semi}, with weights $\Wbf_0^{(l)}\in \Rm^{E\times E}$ and $\Wbf_1^{(l)}\in \Rm^{E\times E}$ and hidden activations $\Hbf^{(l)}\in \Rm^{N\times E}$ at layer $l$, and is given as:
\begin{equation}
\Hbf^{(l+1)} = \mathlarger \sigma\Big( \Hbf^{(l)}\Wbf_0^{(l)}+\Dbf^{-1}\Abf_{\text{I}}\Hbf^{(l)}\Wbf_1^{(l)}\Big),
\label{eq:encoder}
\end{equation}
where $\Dbf$ is the degree matrix obtained from the adjacency matrix with diagonal entries, $D_{ii} = \sum_{j=1}^N A_{ij}$. Note that for the first layer, we set $\Hbf^{(0)}=\Xbf$ and for the last layer $\Hbf^{(L)}=\Zbf$. Also, notice from~\eqref{eq:encoder} that the encoder updates each node with a weighted transformation of node features of its neighbours. A non-linear decoder is used to predict the adjacency matrix from thus encoded embedding: $\alphabf = g(\Zbf)$. 
We use a radial basis function kernel as the decoder operating on each pair of nodes i.e., 
\begin{equation}
\alpha_{ij} = \exp  \big(-\frac{1}{2}(\zbf_i-\zbf_j)^2\big), \quad \forall (i,j).
\label{eq:decoder}
\end{equation}
Position and radius information of adjacent nodes, according to $\alphabf$, are used to generate rough binary segmentations by drawing 3D spheres along each edge.

%Rough binary segmentations are generated from the predicted adjacency matrix, ${\alphabf}$, using position and scale information from node feature matrix, $\Xbf$, for evaluation purposes. \todo{TK: Can we maybe be a little bit more specific?}
%\subsubsection*{Dice loss}

{\bf Dice loss: }
Weights of the GNN model, $\Wbf^{(l)}$, are learned by minimising dice loss between predicted adjacency at the final layer, ${\alphabf}$, and ground truth labels, $\Abf$, of training data. Dice loss is useful to account for class imbalance (between edge and no-edge classes) and is given as:
\begin{equation}
\mathcal{L}(\Abf, {\alphabf}) = 1-\frac{2\sum_{i,j=1}^N {A}_{ij}\alpha_{ij}} {\sum_{i,j=1}^N  {A}_{ij}^2 + \sum_{i,j=1}^N {\alpha}_{ij}^2}.
        \label{eq:dice}
\end{equation}

\section{Experiments and Results}
\label{sec:exp}

%\subsubsection*{Data and Parameters}

We use two disjoint subsets of CT scans of size 32 and 100 subjects from the Danish lung cancer screening trial~\citep{pedersen2009danish}. The first subset, with 32 scans, has expert-verified reference segmentations and is used for comparing the methods, whereas the second subset, of 100 scans, has reference segmentations obtained automatically using the airway extraction method in~\citep{lo2009airway} and it is used
to perform pre-training and tune the hyperparameters. 

\begin{figure*}
\centering
\begin{subfigure}[b]{0.46\textwidth}
\centering
\includegraphics[width=0.99\linewidth]{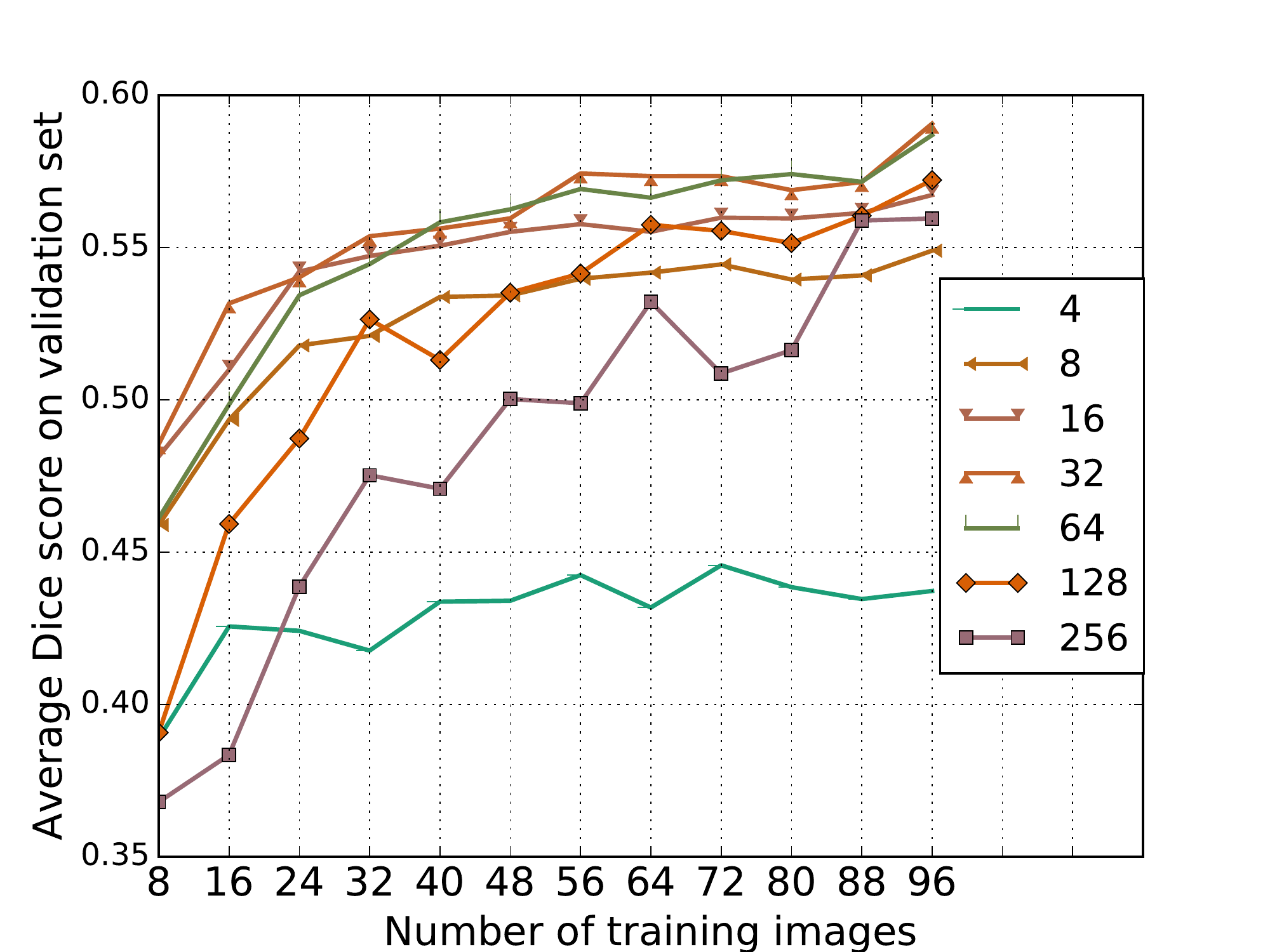}
\vspace{-0.5em}
\caption{Plot showing the influence of increasing training set size on average dice score on the validation set for different hidden units per GNN layer.}
\label{fig:num_train}
\end{subfigure}
\hspace{0.4cm}
\begin{subfigure}[b]{0.46\textwidth}
\centering
\includegraphics[width=0.7\linewidth]{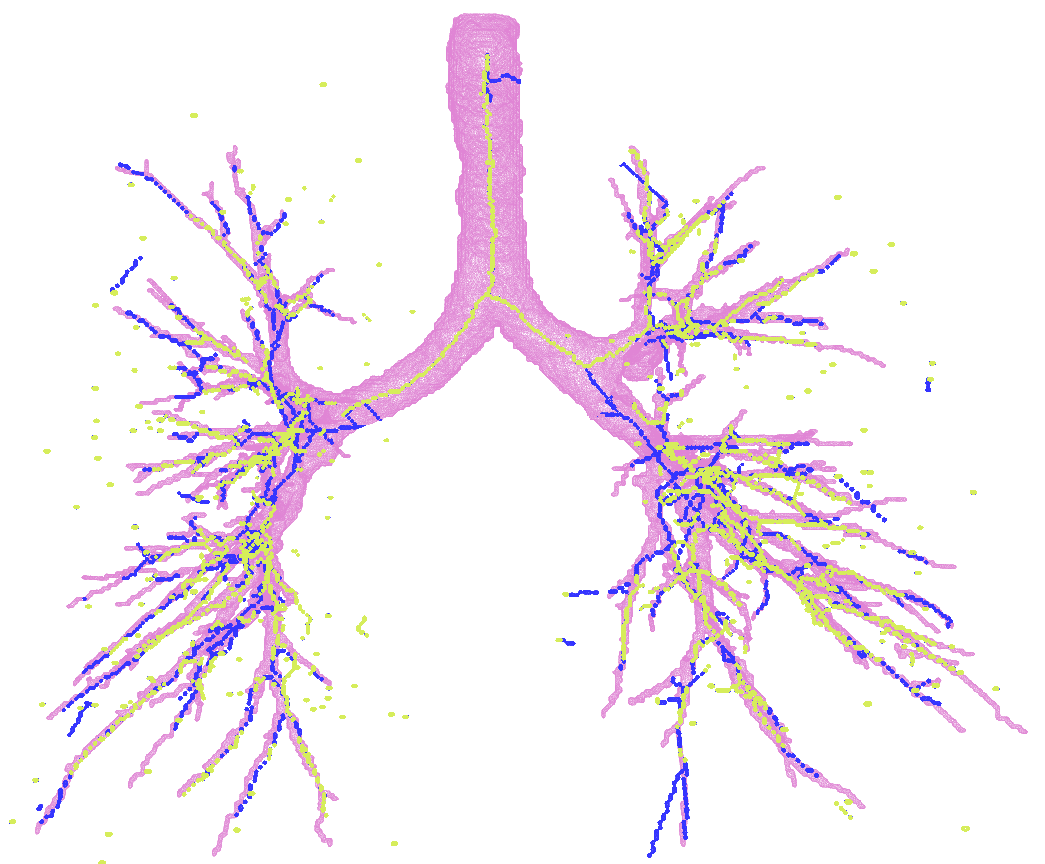}
% \vspace{-0.5em}
\caption{Airway tree centerlines for one of the test cases obtained from MFN predictions (blue) overlaid with the reference segmentations (pink surface) and the
centerlines from GNN model (yellow).}
\label{fig:res}
\end{subfigure}
% \vspace{-0.5em}
\end{figure*}

Tuning of GNN model is performed in three stages: hyperparameter tuning, pre-training and final model training. Using the dice loss in~\eqref{eq:dice} for tuning, we set the number of GNN layers to $3$ and number of hidden units per layer to $32$ (see   Figure~\ref{fig:num_train}). We use Adam optimiser with learning rate $0.005$ and batch size of $4$ for training the GNN model.  Further, we use all $100$ scans with automatic reference segmentations to pre-train the GNN model. For the comparing mean-field network model, we use the hyperparameter setting reported in~\citep{selvan2018mean}.

Centerlines are extracted from the generated binary segmentations to compute centerline distance, which is used to evaluate the performance of the methods. It is defined as $d_{err} = (d_{FP} + d_{FN})/2 $, where $d_{FP}$ is the average minimum Euclidean distance from extracted centerline points to reference centerline points, and $d_{FN}$ is the average minimum Euclidean distance from reference centerline points to extracted centerline points.

%\subsubsection*{Results}
{\bf Results } Experiments comparing GNN and MFN models were conducted on the $32$ reference scans split into $24$ for training and $8$ for test purposes. Table~\ref{tab:res} summarises centerline distance error, $d_{err}$, on the test set. We show two versions of the GNN model: a stand-alone model (GNN) and with merged predictions from MFN (GNN+MFN), obtained from union of the predictions from both methods. Visualisation of MFN and GNN predictions on one test set image along with the reference segmentations is shown in Figure~\ref{fig:res}.

%\begin{wraptable}{l}{0pt}
\begin{table}[h]
%\begin{small}
  \caption{Performance comparison using centerline distance}
\label{tab:res}
  \centering
  \begin{tabular}{lccc}
    \toprule
    %\multicolumn{2}{c}{Part}                   \\
    %\cmidrule{1-2}
Method & $d_{FN}$(mm) & $d_{FP}$(mm) & $d_{err}$ (mm)  \\
    \midrule
%    {Voxel Classifier} & $0.792$ & $4.807$ & $2.799 \pm 0.701 $ \\
%    \small{Bayesian Smoothing} & $0.839$ & $2.812$ & $1.825 \pm 0.232$  \\
    {MFN} &  $2.571$ & $0.835$  & $1.703 \pm 0.186$   \\
    {GNN} &  $2.890$ & $3.913$ & $3.402 \pm 0.386$   \\
    {GNN+MFN} & $2.014$ &  $3.345$  & $2.679 \pm 0.264$   \\
    \bottomrule
  \end{tabular}
  \vspace{-0.5cm}
%  \end{small}
\end{table}
%\end{wraptable}

\section{Discussion}
\label{sec:disc}

The error measures in Table~\ref{tab:res} indicate that as a stand-alone method, GNN model does not compare favourably to MFN model. For the combined case of GNN+MFN we see that there is reduction in the $d_{FN}$ measure indicating that the GNN model is able to add missing branches that are not detected by the MFN model. There are two main reasons, we believe, for the observed performance of GNN model. Firstly, a limited amount of labelled data, which is reflected in Figure~\ref{fig:num_train}, wherein it appears that improvement in accuracy may follow from further increases in training set size. Secondly, the investigated GNN model models only nodes and does not take pairwise node interactions into account. For airway extraction tasks, modelling pairwise interactions can be beneficial~\citep{selvan2018mean}. A variation of GNN which explicitly learns representations for pairs of connected nodes may improve performance. 

As a final remark, with our evaluation we observe that use of hand-crafted models such as MFNs can be useful in cases where domain knowledge is available. However, end-to-end trainable models using GNNs can be useful when there is limited insight into the exact nature of problems and may have greater potential with future growth in dataset sizes. 
%can be  whereas when there is limited insight into the exact nature of problems and larger training data is available formulating end-to-end trainable models such as GNNs may have greater potential.

%Learning node features directly from CT data using dense neural networks, instead of providing high-level features as done in the current model, can be more conducive in end-to-end learning settings. 
%In this work, we explored the possibility of using graph neural networks to perform graph refinement, with an application to extract airway trees from CT data. While results are still not superior to MFN, the more flexible GNN model may have greater potential with future growth in data set sizes. 

%In this work, we explored the possibility of using graph neural networks to perform graph refinement, with an application to extract airway trees from CT data, and our preliminary investigation indicates that the method has potential with interesting future directions to explore: a) use more useful training data b) try out further complex models with dense layers to obtain complex combination of features c) use GNNs that have edge-based nodes to capture node interactions more explicitly.

%\subsubsection*
{\bf Acknowledgements } This work was funded by the Independent Research Fund Denmark (DFF) and the SAP Innovation Center Network.

\bibliographystyle{unsrt}
\medskip

\small
\vspace{-0.3cm}
\setlength{\bibsep}{4pt plus 0.4ex}
\bibliographystyle{IEEEtran}
\bibliography{references.tex}
\end{document}